\documentclass{article}

\PassOptionsToPackage{numbers, compress}{natbib}

\usepackage[main,final]{neurips_2026}

\usepackage[utf8]{inputenc} 
\usepackage[T1]{fontenc}    
\usepackage{hyperref}       
\usepackage{url}            
\usepackage{booktabs}       
\usepackage{amsfonts}       
\usepackage{nicefrac} 
\usepackage{microtype}      
\usepackage{xcolor}         
\usepackage{algorithm}
\usepackage{algpseudocode}
\usepackage{amsmath}
\usepackage{graphicx}
\usepackage{subcaption}
\usepackage{enumitem}
\usepackage{wrapfig}
\usepackage{multirow}
\usepackage{makecell}

\title{Correlation-Aware and Gaussianity-Preserving Robust Latent Angular Watermarking for Diffusion Models}

\author{
  Yebin Zheng \\
  Department of Infocomm Technology \\
  Singapore Institute of Technology \\
  \texttt{yebin.zheng@singaporetech.edu.sg}
  \And
  Haonan An \\
  Department of Computer Science \\
  City University of Hong Kong \\
  \texttt{haonanan2-c@my.cityu.edu.hk}
  \And
  Guang Hua \\
  Department of Infocomm Technology \\
  Singapore Institute of Technology \\
  \texttt{ghua@ieee.org}
  \And
  Zhiping Lin \\
  School of Electrical and Electronic Engineering \\
  Nanyang Technological University \\
  \texttt{ezplin@ntu.edu.sg}
  \And
  Yuguang Fang \\
  Department of Computer Science \\
  City University of Hong Kong \\
  \texttt{my.fang@cityu.edu.hk}
}

\begin{document}
\maketitle

\begin{abstract}
Latent domain watermarking for diffusion models embeds watermarks directly into the latent prior, enjoying non-intrusiveness to model parameters and seamless integration with the generation process. However, due to the violation of latent Gaussianity or sensitivity to normal and malicious perturbations during latent inversion, existing methods are prone to watermark detection or removal attacks. A further overlooked problem is the violation of the i.i.d. latent condition after watermarking, which leads to latent correlation degradation and generation fidelity loss. Although this has been externally measured by FID, the internal correlation structure has yet to be rigorously characterized. To address the above issues, and motivated by the rotation-invariant property of isotropic Gaussian, we propose \emph{Latent Angular Watermarking (LAW)}, which encodes watermark bits as antipodal angles ($\pm\pi/2$ relative to a reference pair) between disjoint pairs of latent elements while preserving the Gaussianity. The antipodal ($\pi$-separation) encoding maximizes geometric separation between bit values, and we prove that the decoding angular-error variance is proportional to the norm of the latent pair, i.e., $\operatorname{var}(\Delta \phi) \propto 1/\rho^2$. We further propose a magnitude-driven variant, LAW-M, which anchors watermark bits in the most geometrically stable latent dimensions, yielding additional robustness gains. Theoretically, we provide a rigorous characterization of the induced correlation degradation, deriving in closed form the autocorrelation structure of the watermarked latent and proving that correlations are confined to a sparse, structured set of off-diagonal elements with fixed $\pm \pi/4$ values. Extensive experiments demonstrate state-of-the-art robustness against a broad range of post-processing and regeneration attacks, together with the best image fidelity among existing methods.
\end{abstract}

\section{Introduction}
\label{sec:intro}
Latent diffusion models (LDMs) \citep{Rombach_2022_CVPR} have revolutionized visual content creation. Yet, their unprecedented ability to synthesize photorealistic images at scale has triggered urgent copyright and provenance concerns. To address them, watermarking has become an essential defense approach. 

Diffusion model watermarking methods are classified as either post-processing or in-generation. While the former embeds watermarks into images after generation, the latter integrates watermarking into the generation process. In-generation methods can be further classified into fine-tuning-based and latent domain embedding, while the latter has gained significant research attention for its training-free nature \citep{wen_2023_nips,ci2024ringid,Yang_2024_CVPR,huang2024robin,Guo_2024_Nips,Lee_2025_ICCV,gunn_2025_undetectable,li2025gaussmarker,hu2026spherical}. Latent domain embedding is thus the focus of this paper.

In-generation watermarks are generally robust against common image post-processing, but they remain vulnerable to dedicated watermark removal attacks \citep{an2026removing,an2026boxfree}. It is found \citep{gunn_2025_undetectable,hu2026spherical} that a binary classifier can be trained to distinguish between watermarked and non-watermarked images, and the classifier can be further used to train a watermark removal network. We attribute this to the disruption of the latent prior Gaussianity after watermarking, creating detectable artifacts in generated images.

Recent studies have explored methods that preserve the Gaussianity \citep{Yang_2024_CVPR, gunn_2025_undetectable, hu2026spherical}, so the latent prior after watermarking still follows a Gaussian distribution. Nevertheless, similar to other prior works, their watermark robustness hinges on repetitive encoding and majority voting \citep{Yang_2024_CVPR, gunn_2025_undetectable, hu2026spherical}, which inevitably decreases generation diversity and still introduces visual artifacts.

Besides robustness, watermark fidelity in the latent domain embedding paradigm has also been understudied. Note that watermarking can induce correlations among latent elements, breaking the i.i.d. condition even when Gaussianity remains unchanged. Despite the existing external measurements such as Fr\'echet Inception Distance (FID)~\citep{heusel2017fid}, the internal correlation degradation in the latent space has not been rigorously characterized, leaving a gap in understanding watermark fidelity.

\begin{wraptable}{r}{0.62\columnwidth}
\vspace{-1.0em}
\centering
\caption{Comparison of Latent Domain Watermarking.}
\label{tab:qualitative_compare}
\scriptsize
\setlength{\tabcolsep}{2pt}
\resizebox{\linewidth}{!}{
\renewcommand{\arraystretch}{1.0}
\begin{tabular}{r|ccccc}
\toprule
Method 
& \makecell{Preserving\\Gaussianity} 
& \makecell{Preserving\\i.i.d.} 
& \makecell{Repeated\\Embedding} 
& \makecell{Per-image\\key}
& \makecell{Watermark\\Capacity} \\
\midrule
Tree-Ring \citep{wen_2023_nips}  
& No & No & No & No & - \\
RingID \citep{ci2024ringid}  
& No & No & No & No & 11 \\
HSTR \citep{Lee_2025_ICCV}  
& No & No & No & No & - \\
HSQR \citep{Lee_2025_ICCV}  
& No & No & No & No & 72 \\
ROBIN \citep{huang2024robin}  
& No & No & No & No & - \\
Gaussian Shading \citep{Yang_2024_CVPR}  
& Yes & Yes & Yes & Yes & 256 \\
PRC \citep{gunn_2025_undetectable}  
& Yes & No & Yes & No & 512 \\
\midrule
LAW 
& Yes & No (Controlled) & Yes & No & 512 \\
LAW-M 
& Yes & No (Controlled) & No & Yes & 4096 \\
\bottomrule
\end{tabular}%
}
\vspace{-1.0em}
\end{wraptable}

To address the above issues, and motivated by the rotation-invariant property of isotropic Gaussian \citep{Box_1958_MS,hu2026spherical}, we propose \emph{Latent Angular Watermarking (LAW)}, which encodes watermark bits as antipodal angles within the latent prior. Specifically, the latent prior is organized into disjoint pairs and partitioned into an encoding region, a reference region, and a non-encoding region, respectively. Then, each bit is embedded by setting the relative angle between an encoding pair and its reference pair to $\pm \pi/2$ via rotation. On the theoretical side, we derive in closed form the autocorrelation matrix of the watermarked latent and prove that the induced correlations are confined to a sparse, structured set of off-diagonal elements with fixed $\pm \pi/4$ values. For high-stakes scenarios such as preventing synthetic public figure images, we additionally propose LAW-M, a magnitude-driven variant that anchors watermark bits in the most geometrically stable latent dimensions, at the cost of a small per-image private key overhead (similar to Gaussian Shading \citep{Yang_2024_CVPR} with a unique key per image).

The qualitative comparison of the proposed angular watermarking against other latent domain watermarking methods is shown in  Table~\ref{tab:qualitative_compare}. Notably, Gaussian Shading \citep{Yang_2024_CVPR} can strictly preserve i.i.d., but it requires a unique key for each image it watermarks. Our method (LAW), on the other hand, does not have the storage overhead and achieves state-of-the-art robustness and capacity, with strictly constrained i.i.d. degradation. 

Our contributions are as follows:

\begin{itemize}[leftmargin=10pt]
    \item We propose LAW, a novel correlation-aware and Gaussianity-preserving watermarking framework for diffusion models that encodes watermark bits as antipodal angles between disjoint latent pairs, achieving strong robustness through maximal angular separation.
    \item We provide a rigorous theoretical treatment of the correlation degradation in Gaussianity-preserving watermarks. By deriving the closed-form autocorrelation matrix of the watermarked latent, we quantitatively characterize the induced i.i.d. degradation.
    \item We further propose LAW-M, a magnitude-driven variant that anchors watermark bits in the most geometrically stable latent dimensions, yielding a provable bound on decoding angular-error variance and significantly improving robustness for high-stakes scenarios.
    \item We conduct extensive experiments against a broad set of post-processing and regeneration attacks, demonstrating that our method achieves state-of-the-art robustness while simultaneously attaining the best image fidelity among existing diffusion model watermarking methods.
\end{itemize}

\section{Related Work}
\subsection{Post-Processing Watermarking}
Post-processing watermarking involves embedding imperceptible watermarks into an image after the generation process. Early research focused on transform-domain techniques. For instance, DwtDct \citep{alhaj2007combined} incorporated discrete wavelet transform (DWT) and discrete cosine transform (DCT), while DwtDctSvd\citep{Navas_2008_icst} further improved robustness with the use of singular value decomposition (SVD). 

In the deep learning regime, HiDDeN \citep{Zhu_2018_ECCV} pioneered an end-to-end encoder-decoder framework utilizing differentiable noise layers to simulate common attacks. Built on this, RivaGAN \citep{Zhang2019RobustIV} leverages a GAN to enhance watermark fidelity, while TrustMark \citep{Bui_2025_ICCV} addresses scalability by designing a framework compatible with arbitrary-resolution images. \citep{zhang2022deep,zhang2024robust}, on the other hand, improved watermark robustness against model stealing and image post-processing attacks. Another branch of work focused on high-frequency components in generated images and resulted in improved fidelity \citep{zhang2024suppressing} and robustness \citep{chen2024highfrequency}, respectively. Alternatively, the RAW framework \citep{wang2024raw} departed from the encoder-decoder paradigm and reformulated watermark detection as binary classification.

Despite the advances, post-processing watermarking is, in essence, image watermarking that is decoupled from the model, and it therefore does not strictly or effectively protect the model itself.

\subsection{In-Generation Watermarking}
In-generation watermarking embeds watermarks directly into the image generation process. Early methods relied on model fine-tuning. For instance, Stable Signature \citep{Fernandez_2023_ICCV} jointly trains a VAE decoder and a watermark decoder, while AquaLoRA \citep{Feng_2024_icml} injects watermark features into the diffusion backbone via LoRA. However, these methods have to alter the protected model parameters and incur substantial training overhead.

Latent domain watermarking emerged as a fine-tuning-free alternative. Tree-Ring \citep{wen_2023_nips} pioneered this by embedding patterns into the latent prior's Fourier domain, followed by improvements such as RingID \citep{ci2024ringid} for geometric robustness and ROBIN \citep{huang2024robin} for jointly optimizing fidelity and robustness through adversarial training. FreqMark \citep{Guo_2024_Nips} further enhances robustness against regeneration attacks through latent frequency optimization. Alternatively, HSTR and HSQR \citep{Lee_2025_ICCV} maintain frequency integrity by leveraging Hermitian symmetry to improve robustness against regeneration removal. 

To preserve Gaussianity, Gaussian Shading \citep{Yang_2024_CVPR} encodes the watermark bits onto the signs of the latent prior, but it requires a unique key per image, resulting in key management overhead. PRC watermarking \citep{gunn_2025_undetectable} achieves Gaussianity through pseudorandom error-correcting codes, but this is at the price of coding overhead and error-correction-bounded robustness. Recently, Spherical Watermarking \citep{hu2026spherical} attempts to preserve Gaussianity via random padding and spherical projection. All the above methods apply repetitive encoding and majority voting, which decreases generation diversity and affects watermark fidelity. Our work extends this line of research.

\subsection{Watermark Removal Attacks}
Since watermarked images are fully accessible to end users, adversaries can work on watermark removal before redistribution, subject to preserving image quality. Normal post-processing operations such as noise addition, blurring, contrast adjustment, JPEG compression, etc., could degrade watermark extraction to different extents.

Beyond post-processing, more advanced removal attacks have also been studied. For example, it has been shown that perturbations to the latent features of watermarked images can remove the watermark \citep{zhao2024invisible}. In the pixel domain, a deep-learning-based erase and repair pipeline has also been shown to be effective \citep{liu2023erase}. More recently, a dedicated watermark removal network is is trained via query-based reverse engineering \citep{an2026removing} as well as proxy models with transferability \citep{an2026boxfree}.

These attacks apply to any image generation model, including diffusion models. Improving robustness against both post-processing and dedicated removal attacks is one objective of this paper.

\section{Method}
\subsection{Threat Model}
We consider a scenario where a model developer deploys a pre-trained diffusion model as a cloud-based service via APIs and embeds an imperceptible watermark into each generated image. We assume an adversary has access to watermarked images and aims to remove the embedded watermark. The adversary is allowed to apply arbitrary image transformations, including post-processing, regeneration via diffusion or VAE, and adversarial perturbations, with the constraint of preserving image quality. We assume the adversary has no access to any watermarking key. 

\begin{figure}[h]
\centering
\includegraphics[width=\linewidth]{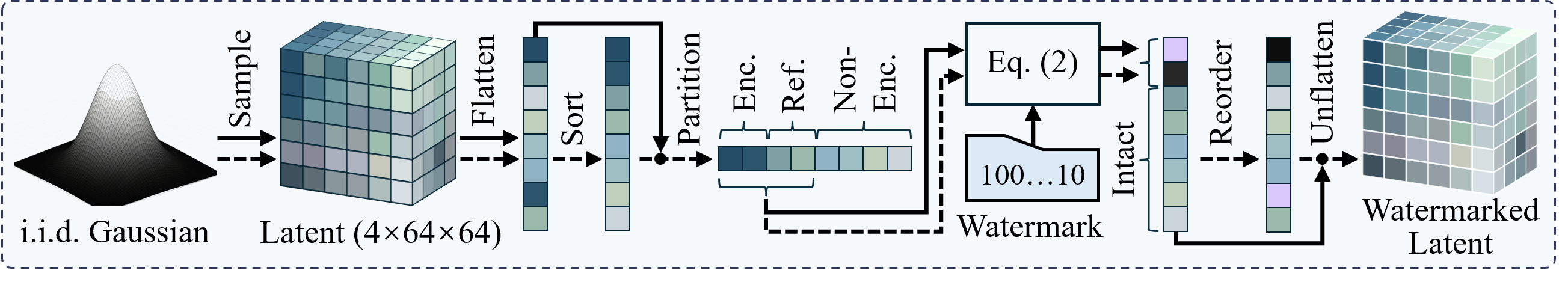}
\caption{The embedding processes of LAW and LAW-M. Solid line: LAW. Dashed line: LAW-M.}
\label{fig:flowchart1}
\end{figure}

\subsection{The LAW Framework}
\begin{wrapfigure}{r}{0.55\textwidth}
\vspace{-10mm}
\centering
\includegraphics[width=\linewidth]{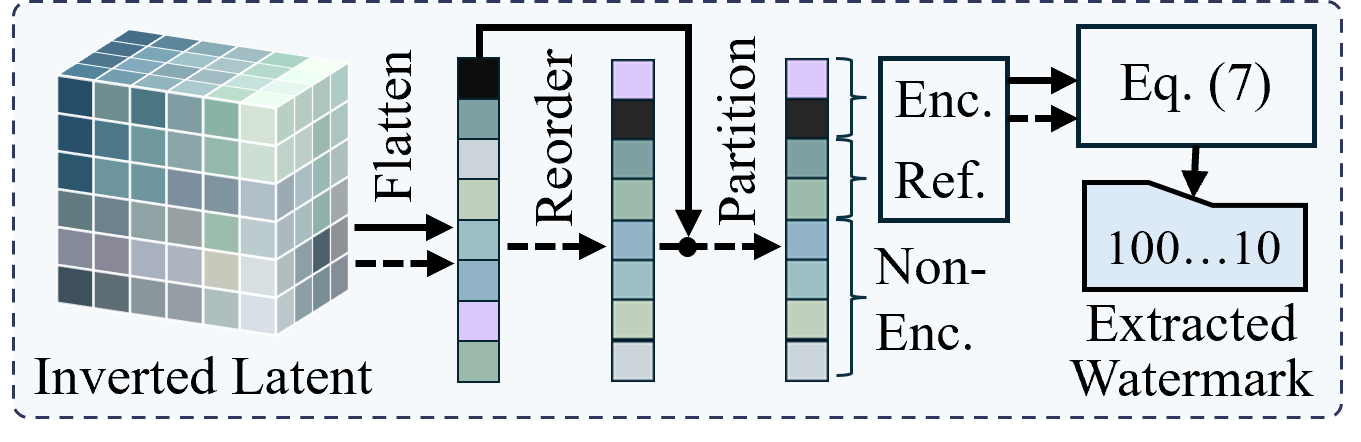}
\caption{The extraction processes of LAW and LAW-M. Solid line: LAW. Dashed line: LAW-M.}
\label{fig:flowchart2}
\vspace{-3mm}
\end{wrapfigure}

LAW is inspired by the rotation-invariant property of isotropic Gaussian \citep{Box_1958_MS,hu2026spherical}, indicating that the rotation of any vectorized subset (a subset of two elements in our setting) of latent prior elements with an arbitrary angle will not affect Gaussianity. This enables us to encode watermark bit $0$ or $1$ into the latent by rotating a vectorized subset to $+ \pi/2$ or $- \pi/2$, with respect to a reference subset, which achieves maximum angular separation for robustness. With this setting, we propose the following angular watermark embedding (Figure \ref{fig:flowchart1} solid flow) and extraction (Figure \ref{fig:flowchart2} solid flow) frameworks.

\subsubsection{Angular Watermark Embedding}
\label{sec:encoding_base}
Let the LDM's initial noise (latent prior) be $z$, containing i.i.d. Gaussian elements, then, we first flatten it into a vector $v \in \mathbb{R}^D$, where $D$ is the total dimension of $z$, e.g., $z = 16,384$ in \citep{Rombach_2022_CVPR}. For an $L$-bit watermark $m \in \{0,1\}^L$, we partition $v$ into $D/2$ disjoint element pairs, denoted by 
\begin{equation}\label{eq:v}
v = [ \underbrace{x_1,x_2,\ldots,x_L}_{2L\text{ Enc.}}, \underbrace{r_1,r_2,\ldots,r_L}_{2L\text{ Ref.}},\underbrace \ldots _{D - 4L\text{ Non-Enc.}}],
\end{equation}
where $x_i = [x_{i,1}, x_{i,2}]\in{\mathbb{R}^2}$ represents encoding pairs and $r_i = [r_{i,1}, r_{i,2}] \in{\mathbb{R}^2}$ represents reference pairs. With this partition, watermark bit $m_i$ can be encoded through rotating $x_i$ with respect to the reference $r_i$. Using the polar coordinates, let $\rho_{x,i} = \|x_{i}\|_2$ and $\phi_{r,i} = \operatorname{atan2}(r_{i,2}, r_{i,1})$, the watermark embedding rule is then given by
\begin{equation}
\label{al:encoding}
\phi'_{x,i} =
\begin{cases}
\phi_{r,i} + \frac{\pi}{2}, & \text{if }m_i = 0, \\
\phi_{r,i} - \frac{\pi}{2}, & \text{if }m_i = 1,
\end{cases}
\end{equation}
and the portion in $v$ that has been altered due to watermark embedding is $x_i$, denoted by
\begin{equation}
x'_{i} =
\left[
\rho_{x,i} \cos \phi'_{x,i},
\rho_{x,i} \sin \phi'_{x,i}
\right],
\end{equation}
which leads to
\begin{equation}\label{eq:v_wm}
v' = [x'_1,x'_2,\ldots,x'_L, r_1,r_2,\ldots,r_L,\ldots],
\end{equation}
which is then reshaped as the watermarked latent and fed into the denoising pipeline, generating the watermarked image denoted by $I_m$. LAW encodes arbitrary watermarks bit by bit, mirroring classic watermarking, unlike learning-based approach \citep{Min_ECCV_2024} requiring supervision from sampled watermark instances. Law also requires no prior knowledge of $z$ and does not incur storage overhead.

\subsubsection{Angular Watermark Extraction}
\label{sec:decoding_base}
The corresponding watermark extraction is generally an inversion of the embedding process. For an image under investigation, denoted by $\hat{I}$, which could be watermarked, non-watermarked, or subject to post-processing or advanced attacks, we first apply DDIM inversion \citep{song2021ddim} to obtain the estimate of the potentially watermarked latent prior that generated the image. After that, we apply the same vectorization (flattening) and partitioning to the inverted latent, yielding the corresponding encoding, reference, and non-encoding regions, i.e.,
\begin{equation}
\hat{v} = [ \hat{x}_1,\hat{x}_2,\ldots,\hat{x}_L,\hat{r}_1,\hat{r}_2,\ldots,\hat{r}_L,\ldots].
\end{equation}
Then, each watermark bit is extracted by comparing the encoding pair with the corresponding reference. To achieve so, we first obtain the phase terms $\hat{\phi}_{x,i} = \operatorname{atan2}(\hat{x}_{i,2}, \hat{x}_{i,1})$, and $\hat{\phi}_{r,i} = \operatorname{atan2}(\hat{r}_{i,2}, \hat{r}_{i,1})$, and the watermark bit decoding rule is given by
\begin{equation}\label{eq:decode}
\hat{m}_{i} =
\begin{cases}
0, & \text{if } \hat{\phi}_{x,i} > \hat{\phi}_{r,i}, \\
1, & \text{otherwise},
\end{cases}
\end{equation}
where $\hat{m}_i$ is the estimated $i$th watermark bit.

\subsection{Variant: The LAW-M Framework}
We design a variant of LAW, i.e., LAW-M, for high-stakes scenarios to protect important single images, such as those of public figures. The motivation is to further enhance watermark robustness, at the price of some key storage overhead. To achieve so, we turn the non-informed LAW embedding into an informed one, making the embedding process dependent on some robustness-related knowledge of $z$, which brings in magnitude-driven sorting and reordering. The corresponding LAW-M embedding and extraction are illustrated in Figures \ref{fig:flowchart1} and \ref{fig:flowchart2} dashed flows, respectively.

Concretely, LAW-M embedding process differs from LAW by adding a sorting process before partitioning, and with a slight abuse of notation, we reuse (\ref{eq:v}) to express the sorted partition $v_\text{M}$,
\begin{equation}
v_\text{M} = [x_1,x_2,\ldots,x_L, r_1,r_2,\ldots,r_L,\ldots],
\end{equation}
with the additional condition on sorted magnitudes
\begin{equation}\label{eq:sort}
\rho_{x,1} \ge \rho_{x,2} \ge \ldots \ge \rho_{x,L} \ge \rho_{r,1} \ge \rho_{r,2} \ge \ldots \ge \rho_{r,L}.
\end{equation}
The underlying rationale behind the arrangement of (\ref{eq:sort}) is that for the same rotation angle, a large-magnitude vector travels longer than a small-magnitude one in Euclidean space, and this maximizes the attacking budget for the adversary to create a wrong decoded bit using (\ref{eq:decode}). See our proof in Appendix \ref{sec:appendix_proof1} for details. Meanwhile, this does not violate Gaussianity thanks to rotation invariance.

The indices of sorted $x_i$ and $r_i$, denoted by key $\mathcal{K}\in \mathbb{R}^{2L}$, serve as a secret key and are stored for mapping back to the latent, which incurs some storage overhead, determined by the watermark length $L$. On the extraction side, LAW-M differs from LAW only by the additional $\mathcal{K}$-based reordering.

\subsection{Theoretical Analysis}
We provide rigorous theoretical analysis on Gaussianity, i.i.d. degradation, and robustness reflected on the angular error induced by attacks. We summarize the following theoretical results with detailed proofs in the corresponding appendices. 

\textbf{Property 1: Gaussianity Preservation.} Let $z_m$ be the watermarked latent, reshaped from $v'$ in (\ref{eq:v_wm}), then each element of $z_m$ preserves zero mean and unit variance. See proof in Appendix \ref{sec:our_mean_variance}.

\textbf{Property 2: i.i.d. Degradation.} The autocorrelation matrix of $z_m$ has the following structure:
\begin{equation}
\mathcal{R}(z_m, z_m) =
\begin{bmatrix}
I_{2L} & A & 0\\
A^\top & I_{2L} & 0\\
0 & 0 & I_{D-4L}
\end{bmatrix},
\end{equation}
where $A= \mathcal{R}([x'_1,x'_2,\ldots,x'_L],[r_1,r_2,\ldots,r_L])$ is block diagonal,
\begin{equation}
A=\operatorname{diag}(A_1,\dots,A_L), \quad A_i=(1-2m_i)
\begin{bmatrix}
0 & -\frac{\pi}{4}\\
\frac{\pi}{4} & 0
\end{bmatrix},
\end{equation}
and $m_i$ is the $i$th watermark bit. See proof in Appendix \ref{sec:our_covar}.

\textbf{Property 3: Angular Error Variance.} Let the latent perturbation during watermark extraction, due to attacks, be $\epsilon\sim\mathcal{N}(0,\sigma^2 I_2)$, and let the angular deviation be $\Delta \phi$. If $\sigma \ll \rho_{x,i}$, then we have 
\begin{equation}
\operatorname{var}(\Delta\phi)\approx \sigma^2/\rho_{x,i}^2 \propto 1/\rho^2_{x,i}.
\end{equation}
See proof in Appendix \ref{sec:appendix_proof1}.

It can be seen from Property 2 that the additional correlations are confined to a sparse, structured set of off-diagonal elements with fixed $\pm \pi/4$ values. It can be seen from Property 3 that prioritizing watermark embedding in high magnitude pairs can reduce angular deviation, which justifies that LAW-M attains the optimal robustness.

\section{Experiment}
\subsection{Experimental settings}
\label{sec:settings}
\textbf{Implementation Details.}
We conduct all experiments using Stable Diffusion v2.1\citep{Rombach_2022_CVPR}, generating $512 \times 512 \times 3$ images with a latent shape of $4 \times 64 \times 64$. For image generation, we use a 50-step DPM-Solver++~\citep{lu2025dpmsolverpp} with guidance scale 7.5. For latent reconstruction, we employ a 50-step DDIM inversion~\citep{song2021ddim} with guidance scale 1.0. To emulate a realistic blind extraction scenario where the original prompt is unknown, DDIM inversion is performed using empty prompts. All experiments are implemented in PyTorch and on a single NVIDIA RTX 5090 GPU. 

\textbf{Repeated Embedding.} For some experiments, we consider $R$ times repeated embedding of the same watermark message, following existing works \citep{Yang_2024_CVPR,gunn_2025_undetectable,hu2026spherical}. Specifically, we set $R=7$ for LAW and $R=1$ for LAW-M, as the robustness of the latter can be seen from the magnitude-driven sorting mechanism, relaxing the need for repeated embedding. Note that $R$ is bounded by watermark length and embedding capacity.

\textbf{Watermark Baselines.}
We compare our approach against two categories of baselines. For the post-processing baselines, we consider DwtDct~\citep{alhaj2007combined}, DwtDctSvd~\citep{Navas_2008_icst}, RivaGAN~\citep{Zhang2019RobustIV} and TrustMark~\citep{Bui_2025_ICCV}. For the in-generation baselines, we focus on latent domain methods, including Tree-Rings \citep{wen_2023_nips}, RingID \citep{ci2024ringid}, HSTR and HSQR \citep{Lee_2025_ICCV}, ROBIN\citep{huang2024robin}, Gaussian Shading \citep{Yang_2024_CVPR},  and PRC \citep{gunn_2025_undetectable}. Following their source code, the watermark length is set to 32 bits for DwtDct, DwtDctSvd, and RivaGAN, 100 bits for TrustMark, and 512 bits for all other methods.

\textbf{Datasets \& Evaluation Metrics.}
We randomly sample 500 captions from the MS COCO 2017 validation set \citep{lin2014coco} as text prompts for Stable Diffusion. Our evaluation considers two primary criteria, i.e., fidelity and robustness. Image fidelity is evaluated using the FID~\citep{heusel2017fid}, CLIP Score~\citep{radford2021clip}, and Inception Score~\citep{salimans2016improved}. Watermark robustness is measured by bit-wise extraction accuracy (ACC) and the true positive rate at $1\%$ false positive rate (TPR@1\%FPR), hereafter referred to as TPR.

\begin{figure}[t]
    \centering
    \includegraphics[width=\textwidth]{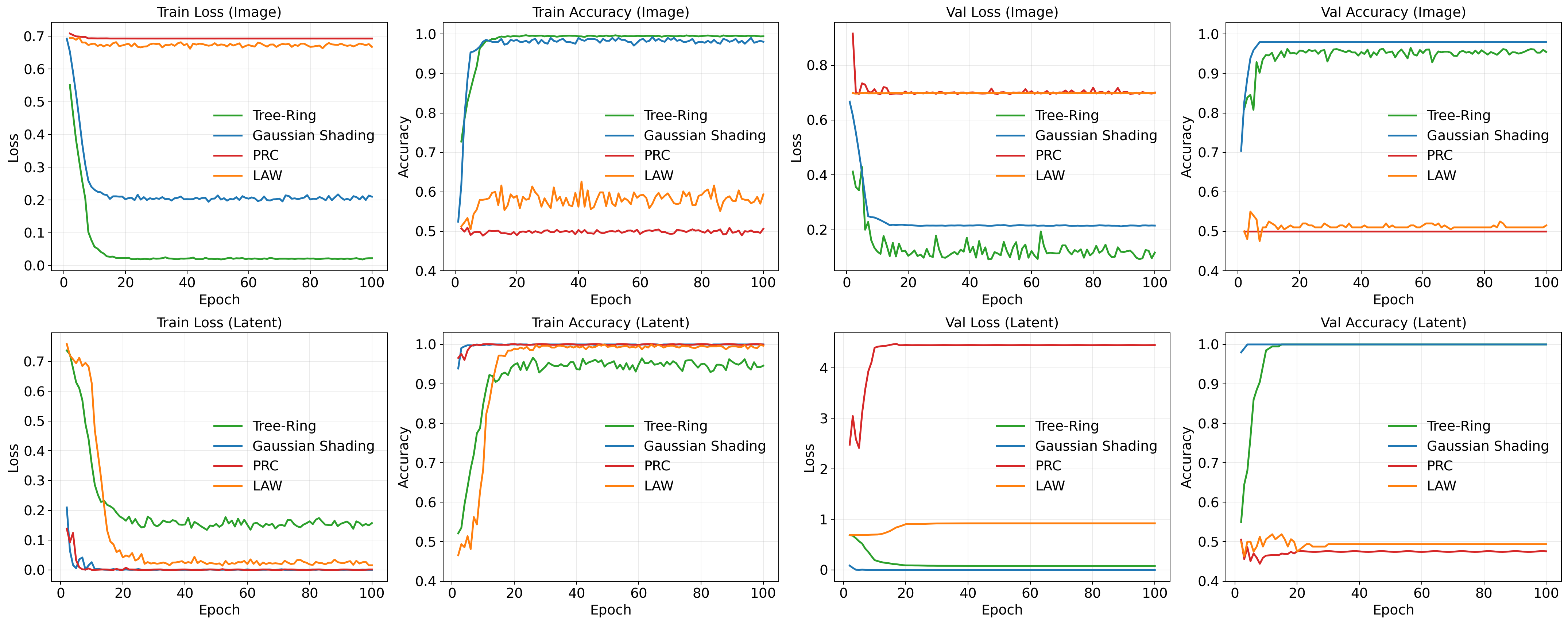}
    \caption{Binary classification (watermark detection) accuracy and loss at image and latent levels.}
    \label{fig:train_loss}
\end{figure}

\subsection{Main Results}
\textbf{Computational Indistinguishability.} A core advantage of preserving Gaussianity is to evade learning-based watermark detection, which is often the first step in dedicated watermark removal attacks. To validate this, we assess the computational indistinguishability of our method at both the image and latent levels (see Figure~\ref{fig:train_loss}). At the image level, we train a ResNet-18 binary classifier~\citep{He_2016_CVPR} to distinguish between watermarked and non-watermarked images. Notably, only PRC and our method successfully maintain a classification accuracy of $\sim 50\%$. At the latent level, a two-layer MLP \citep{Rumelhart1986} fails to differentiate our watermarked latent $z_{m}$ from pristine i.i.d. Gaussian noise $z$. These empirical results well align with our theoretical analysis, confirming that the proposed LAW achieves strong Gaussian indistinguishability. Further experimental details are provided in Appendix~\ref{sec:appendix_exp_details}.

\begin{table}[t]
\centering
\caption{Quantitative fidelity and robustness comparison results. Best: bold. Second best: underscore.}
\label{tab:combined_results}
\small
\resizebox{\linewidth}{!}{
\begin{tabular}{r|ccc|cccccccc|cccc}
\toprule
\multirow{2}{*}{Method} 
& \multicolumn{3}{c|}{Image Quality} 
& \multicolumn{8}{c|}{Post-processing Attacks} 
& \multicolumn{4}{c}{Regeneration Attacks} \\
\cmidrule(lr){2-4} \cmidrule(lr){5-12} \cmidrule(lr){13-16}
& FID$\downarrow$ & CLIP$\uparrow$ & IS $\uparrow$
& Original & Noise & Brightness & Drop & Blur & JPEG & Median & Resize
& VAE-B & VAE-C & Diff & EBRA \\
\midrule

DwtDct \citep{alhaj2007combined}
& $149.3$ & - & $9.6$ & $90.2$ & $21.5$ & $0.0$ & $48.3$ & $0.0$ & $0.0$ & $0.0$ & $58.5$ & $0.0$ & $0.0$ & $0.0$ & $23.2$ \\
DwtDctSvd \citep{Navas_2008_icst}
& $150.0$ & - & $9.7$ & $\mathbf{100.0}$ & $99.5$ & $6.3$ & $65.4$ & $87.3$ & $99.5$ & $98.1$ & $99.5$ & $0.0$ & $0.0$ & $0.0$ & $81.0$ \\
RivaGan \citep{Zhang2019RobustIV}
& $148.3$ & - & $9.5$ & $\underline{97.1}$ & $83.0$ & $56.6$ & $51.2$ & $94.6$ & $59.5$ & $94.6$ & $96.6$ & $0.0$ & $0.0$ & $0.0$ & $73.0$ \\
TrustMark \citep{Bui_2025_ICCV}
& $150.0$ & - & $9.7$ & $95.1$ & $6.8$ & $2.4$ & $72.7$ & $97.6$ & $8.8$ & $95.6$ & $94.6$ & $1.0$ & $0.0$ & $0.0$ & $79.8$ \\
\midrule
Tree-Ring \citep{wen_2023_nips}
& $100.0$ & $\underline{31.3}$ & $14.0$ & $96.2$ & $66.6$ & $75.4$ & $23.4$ & $86.2$ & $81.2$ & $83.8$ & $91.6$ & $67.8$ & $66.2$ & $17.2$ & $32.2$ \\
RingID \citep{ci2024ringid}
& $101.6$ & $31.2$ & $\mathbf{14.7}$ & $\mathbf{100.0}$ & $99.4$ & $98.6$ & $88.0$ & $\mathbf{100.0}$ & $99.6$ & $\mathbf{100.0}$ & $\mathbf{100.0}$ & $\underline{97.2}$ & $98.0$ & $93.6$ & $93.2$ \\
HSTR \citep{Lee_2025_ICCV}
& $99.5$ & $\mathbf{31.4}$ & $\underline{14.5}$ & $\mathbf{100.0}$ & $85.8$ & $94.0$ & $29.8$ & $\underline{99.6}$ & $96.6$ & $98.6$ & $\mathbf{100.0}$ & $82.2$ & $84.6$ & $35.2$ & $45.8$ \\
HSQR \citep{Lee_2025_ICCV}
& $98.5$ & $\mathbf{31.4}$ & $\mathbf{14.7}$ & $\mathbf{100.0}$ & $\mathbf{99.8}$ & $\mathbf{100.0}$ & $86.6$ & $\mathbf{100.0}$ & $\mathbf{100.0}$ & $\mathbf{100.0}$ & $\mathbf{100.0}$ & $\mathbf{97.6}$ & $\mathbf{99.4}$ & $\mathbf{96.8}$ & $92.2$ \\
ROBIN \citep{huang2024robin}
& $99.6$ & $\mathbf{31.4}$ & $14.2$ & $\mathbf{100.0}$ & $99.0$ & $99.0$ & $\mathbf{97.6}$ & $99.0$ & $99.0$ & $99.0$ & $99.0$ & $95.8$ & $97.0$ & $\underline{96.4}$ & $98.0$ \\
\midrule
Gaussian Shading \citep{Yang_2024_CVPR}
& $\underline{97.3}$ & $\mathbf{31.4}$ & $\mathbf{14.7}$ & $\mathbf{100.0}$ & $\underline{99.6}$ & $\mathbf{100.0}$ & $\underline{92.4}$ & $99.3$ & $\underline{99.8}$ & $\underline{99.8}$ & $\mathbf{100.0}$ & $97.0$ & $97.1$ & $35.0$ & $\mathbf{99.0}$ \\
PRC \citep{gunn_2025_undetectable}
& $97.7$ & $\mathbf{31.4}$ & $14.1$ & $\mathbf{100.0}$ & $68.0$ & $75.0$ & $16.0$ & $85.0$ & $85.0$ & $86.0$ & $97.0$ & $63.6$ & $63.4$ & $11.2$ & $21.8$ \\
\midrule
LAW
& $97.4$ & $\underline{31.3}$ & $14.3$ & $\mathbf{100.0}$ & $98.2$ & $97.8$ & $62.2$ & $\underline{99.6}$ & $\underline{99.8}$ & $\underline{99.8}$ & $\underline{99.8}$ & $95.2$ & $94.2$ & $85.0$ & $95.0$ \\
LAW-M
& $\mathbf{96.6}$ & $\mathbf{31.4}$ & $14.3$ & $\mathbf{100.0}$ & $98.8$ & $\underline{99.6}$ & $68.6$ & $\mathbf{100.0}$ & $99.2$ & $\mathbf{100.0}$ & $\mathbf{100.0}$ & $96.4$ & $\underline{98.2}$ & $\underline{96.4}$ & $\underline{98.4}$ \\

\bottomrule
\vspace{-3mm}
\end{tabular}
}
\end{table}

\begin{figure}[t]
    \centering
    \includegraphics[width=\textwidth]{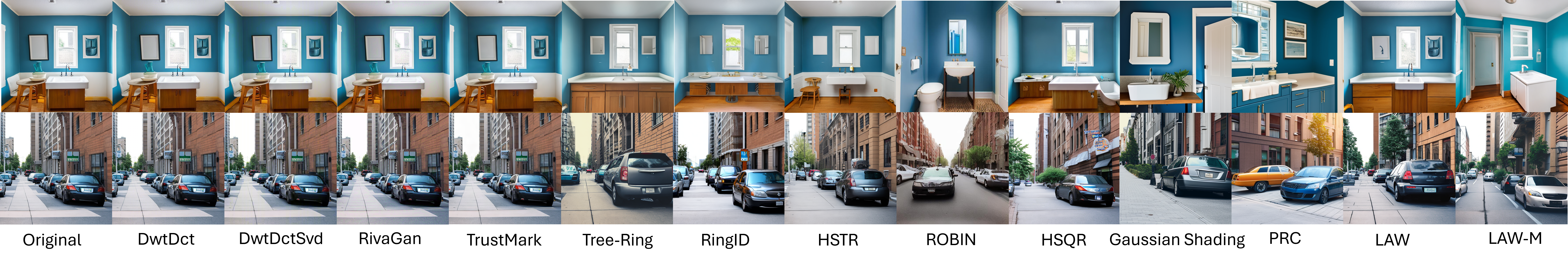}
    \caption{Qualitative comparison of image quality across various watermarking methods.}
    \vspace{-6mm}
    \label{fig:image_quality}
\end{figure}

\textbf{Image Quality and Robustness.} Table~\ref{tab:combined_results} illustrates the trade-off between watermark fidelity and robustness across various methods. In terms of image quality, LAW-M achieves the lowest FID ($96.6$) and a state-of-the-art CLIP score ($31.4$), while maintaining an Inception Score ($14.3$) that is highly competitive with existing latent-domain methods. Crucially, the improved FID of LAW-M over LAW suggests that minimizing latent correlations by avoiding repetitive encoding ($R=1$) better preserves the global image distribution. This qualitative performance is corroborated by the visual comparisons in Figure~\ref{fig:image_quality}, which confirm that our approach maintains high fidelity. Beyond high fidelity, our method demonstrates strong resistance to various attacks. We evaluate watermark extraction accuracy against diverse post-processing and generation-based removal methods (detailed in Appendix~\ref{sec:pr_settings}). As also shown in Table~\ref{tab:combined_results}, our method consistently outperforms state-of-the-art methods across most scenarios, exhibiting particular resilience to severe Gaussian noise, blurring, and deep regeneration attacks. Although extraction accuracy under random dropping falls slightly behind specific baselines like ROBIN, our method achieves superior overall performance, attributed to antipodal encoding and magnitude-driven watermark embedding (LAW-M).

\begin{wrapfigure}{r}{0.5\textwidth}
    \vspace{-3mm}
    \centering
    \includegraphics[width=\linewidth]{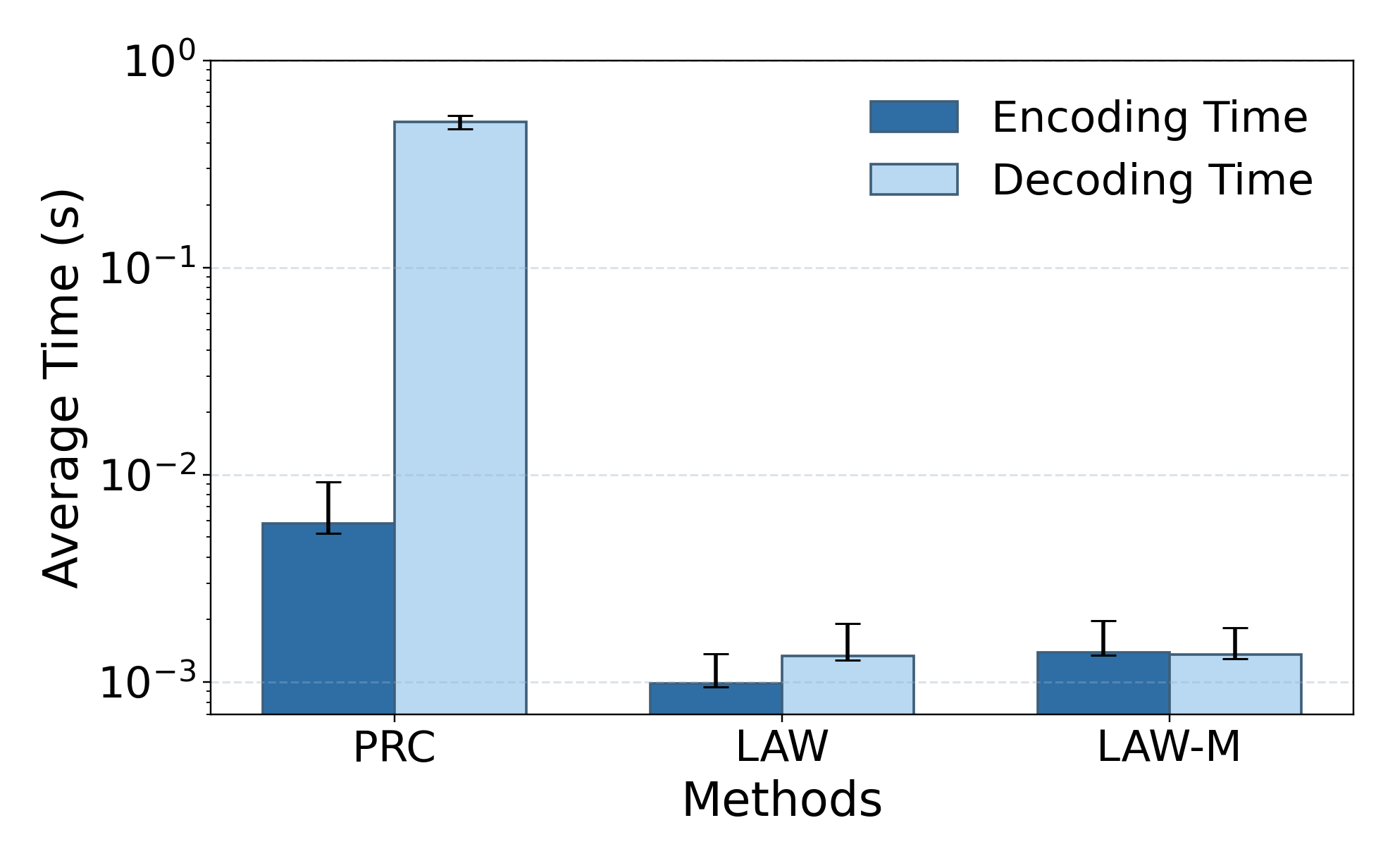}
    \vspace{-5mm}
    \caption{Encoding and decoding time.}
    \label{fig:timing}
    \vspace{-2mm}
\end{wrapfigure}

\begin{figure}[t]
    \centering
    \includegraphics[width=\textwidth]{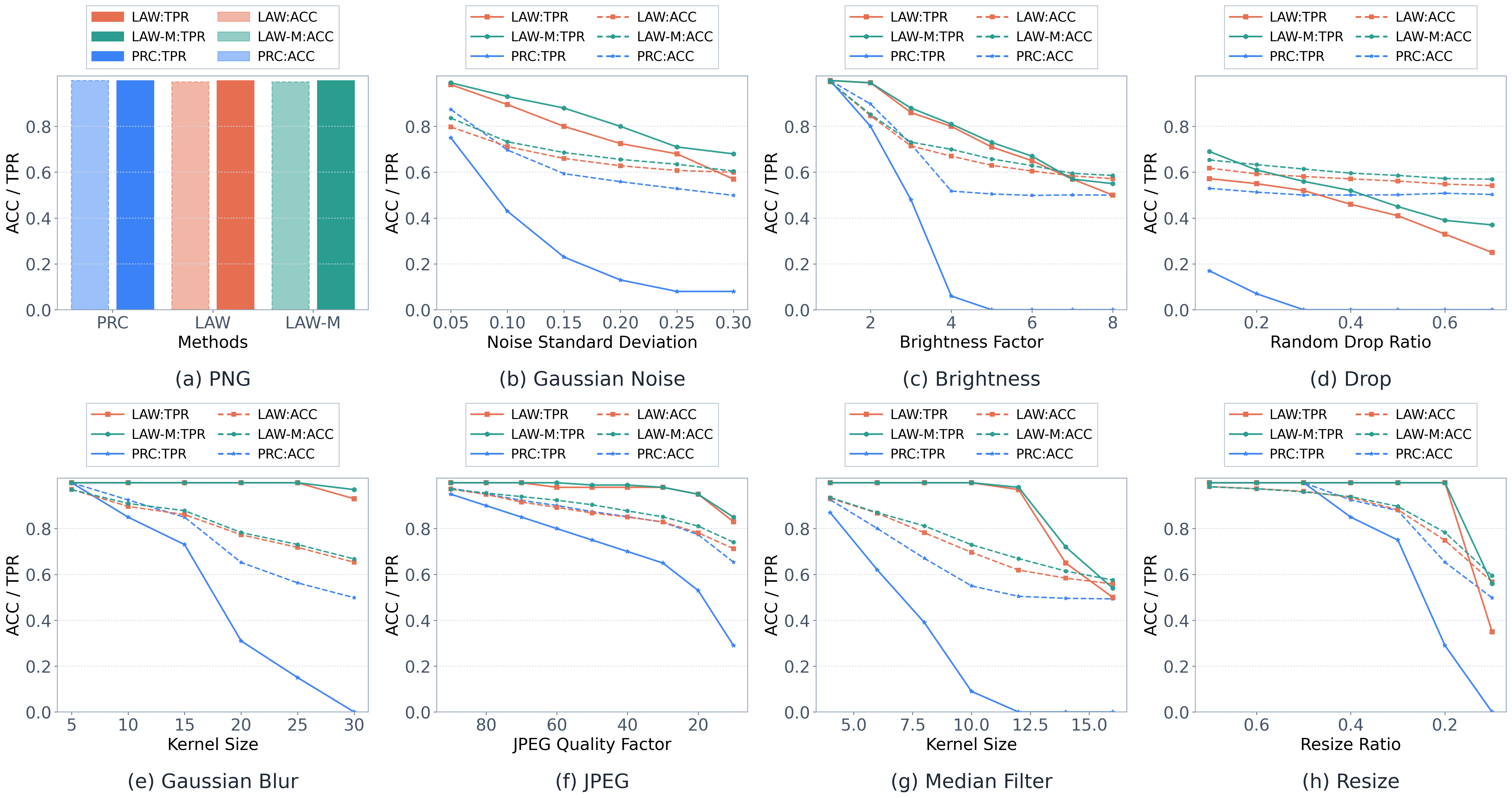}
    \caption{Comparison of PRC watermarking and our method under post-processing attacks.}
    \label{fig:prc_with_ours}
    \vspace{-0mm}
\end{figure}

\textbf{Comparison with PRC Watermarking.} We provide a detailed performance comparison between our method and PRC watermarking under various post-processing attacks, as illustrated in Figure~\ref{fig:prc_with_ours}. The results indicate that our method consistently achieves superior TPR and ACC. Notably, this performance gap becomes more pronounced as the distortion intensity increases. The performance disparity stems from the fundamental difference in encoding mechanisms. While the PRC method relies on pseudo-random error-correcting codes, its error-correction capability is inherently bounded by a hard threshold. In contrast, our method embeds bits such that their representations are strictly separated by $\pi$ in the angular space. Combined with the magnitude-driven anchoring strategy, our method secures a much wider decision margin and achieves significantly higher robustness against severe perturbations.

Furthermore, we evaluate the computational efficiency of both methods, with the average execution time results summarized in Figure~\ref{fig:timing}. Due to the substantial cryptographic overhead, PRC's encoding and decoding latencies are respectively one and three orders of magnitude higher than ours. This significant performance gap demonstrates that our methods are more practical and suitable for real-time deployment in high-throughput generation services.

\begin{figure}[t]
    \centering
    \includegraphics[width=\textwidth]{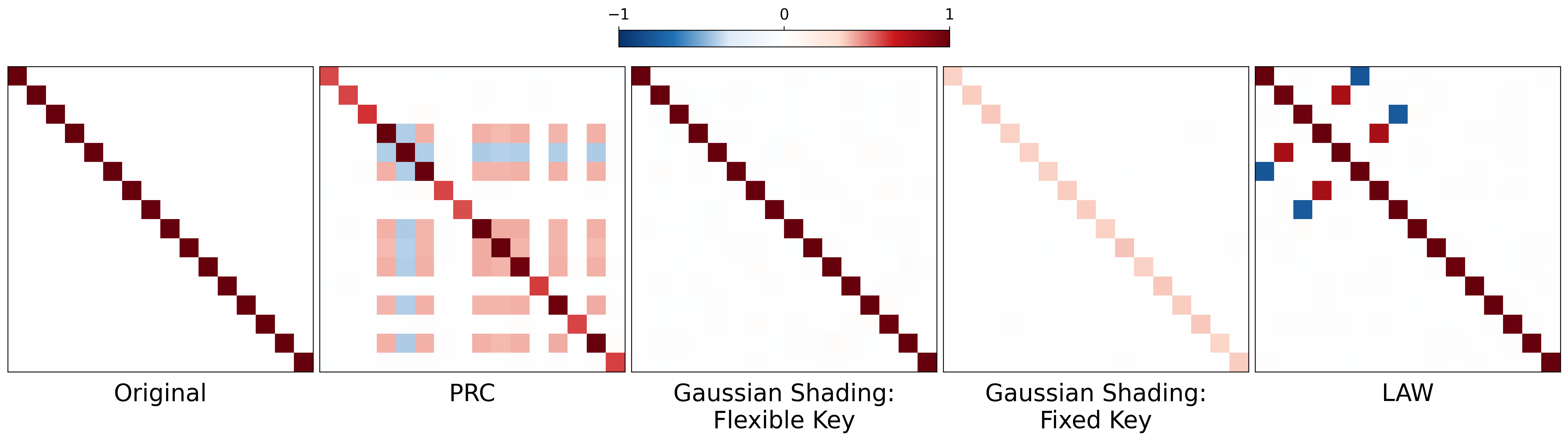}
    \caption{Examples of $16 \times 16$ Autocorrelation matrices of PRC, Gaussian Shading, and LAW.}
    \label{fig:covariance_matric}
\end{figure}

\textbf{Correlation Analysis.} We further evaluate the autocorrelation matrices of three existing Gaussian-mapping methods, with detailed experimental settings provided in Appendix~\ref{sec:covariance}. As illustrated in Figure~\ref{fig:covariance_matric}, Gaussian Shading with a flexible key introduces no correlations. However, this ideal statistical property requires a unique key and nonce for each generated image, leading to significant storage and key management overhead. In contrast, Gaussian Shading with a fixed key alters the variance of the latent distribution, causing a discrepancy with the predefined noise schedule. Consequently, the UNet noise prediction becomes inherently biased at each denoising step, leading to cumulative errors and noticeable degradation in generation quality. Compared to PRC, our method introduces significantly less correlation, thereby achieving superior image quality and higher bit accuracy during watermark extraction. Further analysis on the impact of covariance matrices and the formal mathematical proof regarding the Gaussianity of Gaussian Shading are provided in Appendices~\ref{sec:appendix_cov_analysis} and \ref{sec:appendix_gs_analysis}, respectively.

\subsection{Ablation Study}
\begin{figure}[t]
    \centering
    \includegraphics[width=\textwidth]{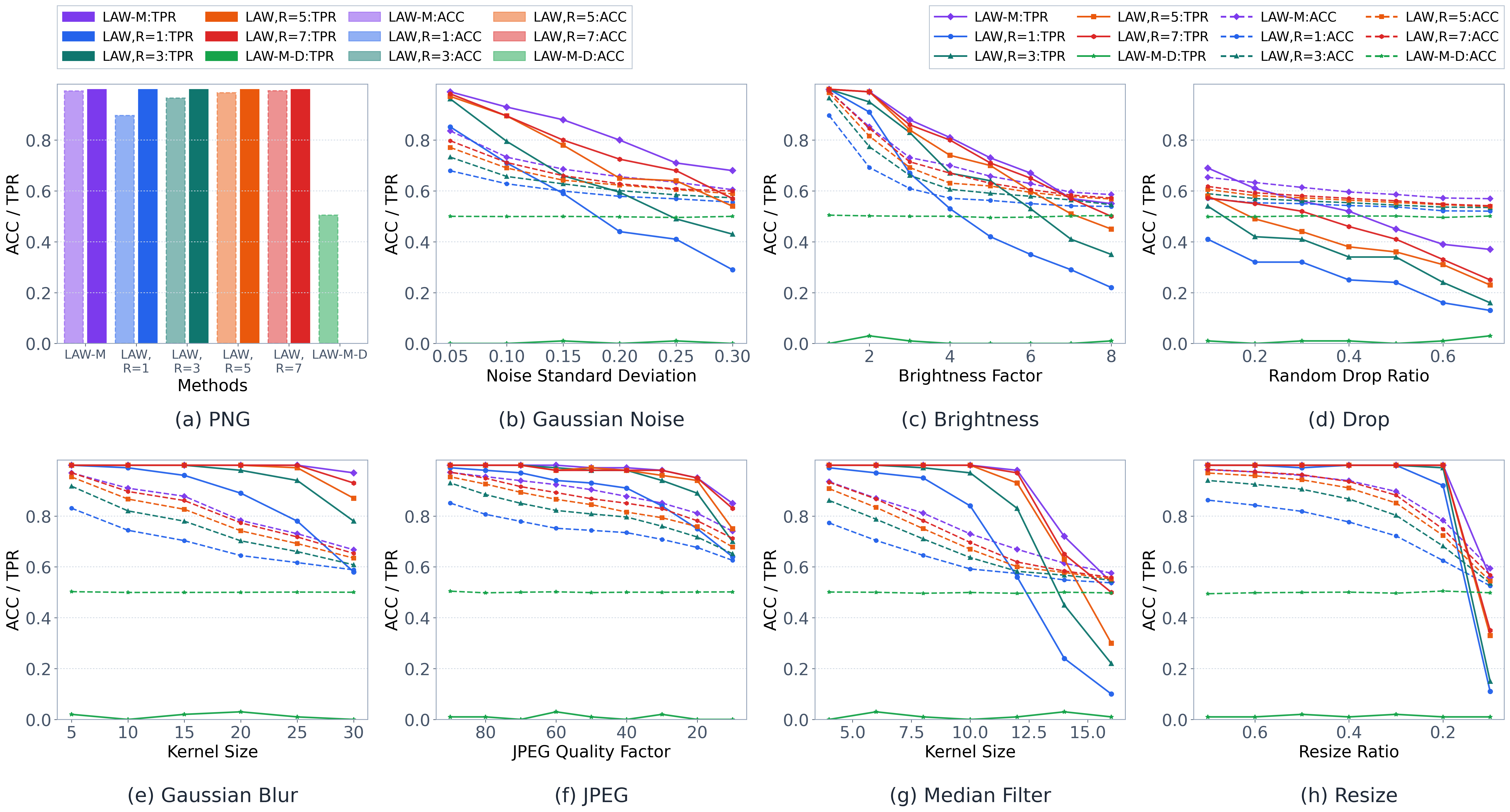}
    \caption{Quantitative results of module ablation and parametric ablation, where LAW-M-D denotes the magnitude-driven variant with decoding based on magnitude sorting at decoding time.}
    \label{fig:ablation_study}
    \vspace{-6mm}
\end{figure}

\textbf{Module Ablation.} We compare the standard LAW with its magnitude-driven variant to evaluate the impact of the magnitude-sorting module. As shown in Figure~\ref{fig:ablation_study}, the magnitude-driven approach achieves superior robustness by anchoring watermark bits in the most geometrically stable latent pairs. However, this gain comes at the cost of storing a per-image private key to preserve the sorting order during extraction.

We further evaluate the necessity of the per-image private key by attempting to extract the watermark using the magnitude order at decoding time. As illustrated in Figure~\ref{fig:ablation_study}, extraction accuracy collapses to nearly zero without the correct key, as even minor perturbations can disrupt the relative magnitude relationships. This resulting misalignment between the encoding and decoding sequences underscores that maintaining the original sorting order via a private key is essential for the robustness of the magnitude-driven variant.

\textbf{Parameter Ablation.} We further evaluate the impact of the repetition factor $R$. As illustrated in Figure~\ref{fig:ablation_study}, increasing $R$ consistently enhances robustness across all post-processing attacks, as redundant embedding significantly mitigates the impact of local perturbations. However, this gain in robustness entails a trade-off with the statistical independence of the latent dimensions. A higher $R$ introduces stronger off-diagonal correlations within the autocorrelation matrix, causing the watermarked distribution to deviate further from the ideal i.i.d. Gaussian distribution.

\section{Limitations}
\label{sec:limitations}
Although LAW and LAW-M achieve strong robustness against most post-processing and regeneration attacks, they remain relatively vulnerable to random drop. In addition, LAW-M requires storing an image-specific private key for magnitude-based latent sorting. While this improves robustness, it also introduces additional storage and key-management overhead.

\section{Conclusion}
We have proposed LAW, a novel diffusion model watermarking framework that exploits the rotation-invariant property of the latent prior. We show that LAW preserves Gaussianity and also improves watermark robustness thanks to the antipodal encoding through rotation, with theoretical evidence. We have also provided a rigorous theoretical analysis of the violation of the i.i.d. condition due to watermarking, despite the preserved Gaussianity, showing that LAW-induced correlations are confined to a sparse, structured set of off-diagonal elements with fixed $\pm \pi/4$ values. A variant, LAW-M, is further developed to protect high-stakes single images, with some storage overhead for the associated secret key. Extensive comparative experiments verified the advantages of the proposed methods. Future work will focus on minimizing or eliminating the induced correlations to achieve strict non-intrusiveness to the latent prior.

\bibliographystyle{plainnat}
\bibliography{ref}

\appendix

\section{Gaussianity Analysis}
\label{sec:gaussianity_analysis}

We analyze the statistical properties of the watermarked latent $z_{m}$ induced by the proposed \textit{Latent Angular Watermarking}. For analytical simplicity, we assume a repetition factor of $R=1$. Since reshaping does not affect the statistical properties, we analyze the vectorized watermarked latent $v'=\operatorname{flatten}(z_{m})$. We first prove that each individual component of $v'$ preserves the zero mean and unit variance. Building upon this, we further characterize the covariance structure of $v'$.

\subsection{Statistical Formulation}

Following Sec.~\ref{sec:encoding_base} and Sec.~\ref{sec:decoding_base}, the vectorized latent is divided into three regions: the encoding region $v'_{enc}$, the reference region $v_{ref}$, and the non-encoding region $v_{non}$. Specifically,
\begin{equation}
v' = \operatorname{concat}(v'_{enc}, v_{ref}, v_{non}),
\end{equation}
where $v'_{enc}=[x_1', \dots, x_L']$, $v_{ref}=[r_1, \dots, r_L]$, and each pair is two-dimensional:
\begin{equation}
x_i'=[x_{i,1}',x_{i,2}'], \qquad r_i=[r_{i,1},r_{i,2}].
\end{equation}
The non-encoding region $v_{non}$ contains all remaining unmodified latent pairs.

For each encoding pair, let $\rho_{x,i}=\|x_i\|_2$ and $\phi_{r,i}=\operatorname{atan2}(r_{i,2},r_{i,1})$. According to the encoding rule(Eq.~\ref{al:encoding}), the modulated encoding pair can be written compactly as
\begin{equation}
x_i'
=
\left[
-(1-2m_i)\,\rho_{x,i}\sin\phi_{r,i},
\ (1-2m_i)\,\rho_{x,i}\cos\phi_{r,i}
\right],
\label{eq:compact_encoding}
\end{equation}
where $m_i \in \{0,1\}$ denotes the $i$-th watermark bit. 

\textbf{Proof.}
We now prove Eq.~\ref{eq:compact_encoding}. If $m_i=0$, then $\phi'_{x,i}=\phi_{r,i}+\frac{\pi}{2}$, and
\begin{equation}
x_i'
=
\left[
\rho_{x,i}\cos(\phi_{r,i}+\tfrac{\pi}{2}),
\rho_{x,i}\sin(\phi_{r,i}+\tfrac{\pi}{2})
\right]
=
\left[
-\rho_{x,i}\sin\phi_{r,i},
\ \rho_{x,i}\cos\phi_{r,i}
\right].
\end{equation}

If $m_i=1$, then $\phi'_{x,i}=\phi_{r,i}-\frac{\pi}{2}$, and
\begin{equation}
x_i'
=
\left[
\rho_{x,i}\cos(\phi_{r,i}-\tfrac{\pi}{2}),
\rho_{x,i}\sin(\phi_{r,i}-\tfrac{\pi}{2})
\right]
=
\left[
\rho_{x,i}\sin\phi_{r,i},
-\rho_{x,i}\cos\phi_{r,i}
\right].
\end{equation}

Noting that $(1-2m_i)=1$ when $m_i=0$ and $(1-2m_i)=-1$ when $m_i=1$, both cases can be written in the unified form of Eq.~\ref{eq:compact_encoding}.
\hfill

\subsection{Mean and Variance}
\label{sec:our_mean_variance}

\paragraph{Property 1. Zero Mean.}
Each dimension of $v'$ has zero mean:
\begin{equation}
\label{al:mean}
\mathbb{E}[v'_j]=0,
\qquad j=1,\dots,D.
\end{equation}

\textbf{Proof.}
The reference region $v_{ref}$ and the non-encoding region $v_{non}$ are not modified, so their means remain zero because the original latent satisfies $z\sim\mathcal{N}(0,I_D)$.

It remains to consider the modified encoding pair $x_i'$. Since the original encoding pair $x_i$ is independent of the reference pair $r_i$, $\rho_{x,i}$ is independent of $\phi_{r,i}$. Moreover, because $r_i\sim\mathcal{N}(0,I_2)$, its phase satisfies
\begin{equation}
\phi_{r,i}\sim \operatorname{Unif}(0,2\pi).
\end{equation}
Therefore,
\begin{equation}
\mathbb{E}[\sin\phi_{r,i}]=0,
\qquad
\mathbb{E}[\cos\phi_{r,i}]=0.
\end{equation}

Using Eq.~\ref{eq:compact_encoding}, we obtain
\begin{equation}
\mathbb{E}[x_{i,1}']
=
-(1-2m_i)\,\mathbb{E}[\rho_{x,i}]
\mathbb{E}[\sin\phi_{r,i}]
=0,
\end{equation}
and
\begin{equation}
\mathbb{E}[x_{i,2}']
=
(1-2m_i)\,\mathbb{E}[\rho_{x,i}]
\mathbb{E}[\cos\phi_{r,i}]
=0.
\end{equation}

Thus, all encoding dimensions also have zero mean. Hence,
\begin{equation}
\mathbb{E}[v']=0,
\end{equation}
which holds for all dimensions in $v'$.
\hfill

\paragraph{Property 2. Unit Variance.}
Each dimension of $v'$ has unit variance:
\begin{equation}
\label{al:variance}
\operatorname{Var}(v'_j)=1,
\qquad j=1,\dots,D.
\end{equation}

Equivalently,
\begin{equation}
\operatorname{diag}\!\left(\operatorname{Cov}(v')\right)=\mathbf{1}_D.
\end{equation}

\textbf{Proof.}
The reference and non-encoding regions are unchanged, so their dimensions still have unit variance. For the encoding region, since $x_i\sim\mathcal{N}(0,I_2)$, we have
\begin{equation}
\mathbb{E}[\rho_{x,i}^2]
=
\mathbb{E}[x_{i,1}^2+x_{i,2}^2]
=
2.
\end{equation}

From Eq.~\ref{eq:compact_encoding} and the zero-mean property,
\begin{equation}
\operatorname{Var}(x_{i,1}')
=
\mathbb{E}[(x_{i,1}')^2]
=
(1-2m_i)^2
\mathbb{E}[\rho_{x,i}^2]
\mathbb{E}[\sin^2\phi_{r,i}]
=
2 \cdot \frac{1}{2}
=
1,
\end{equation}
and similarly,
\begin{equation}
\operatorname{Var}(x_{i,2}')
=
\mathbb{E}[(x_{i,2}')^2]
=
(1-2m_i)^2
\mathbb{E}[\rho_{x,i}^2]
\mathbb{E}[\cos^2\phi_{r,i}]
=
2 \cdot \frac{1}{2}
=
1,
\end{equation}
since $(1-2m_i)=\pm 1$ for $m_i\in\{0,1\}$.
Therefore, every dimension of $v'$ has unit variance, and
\begin{equation}
\operatorname{diag}\!\left(\operatorname{Cov}(v')\right)=\mathbf{1}_D.
\end{equation}
\hfill

\subsection{Covariance Matrix}
\label{sec:our_covar}

\paragraph{Theorem 1. Independence under Separate Transformations.}
Let $U$ and $V$ be independent random variables. For any measurable functions $f$ and $g$, the transformed variables $f(U)$ and $g(V)$ are also independent:
\begin{equation}
\mathbb{E}\!\left[f(U)g(V)\right]
=
\mathbb{E}\!\left[f(U)\right]
\mathbb{E}\!\left[g(V)\right].
\end{equation}
This directly implies
\begin{equation}
\operatorname{Cov}(f(U),g(V))=0.
\end{equation}

\paragraph{Property 3. Internal Covariance within Each Region.}
The covariance matrices within the reference, encoding, and non-encoding regions are identity matrices:
\begin{equation}
\operatorname{Cov}(v_{ref},v_{ref})=I_{2L},
\qquad
\operatorname{Cov}(v'_{enc},v'_{enc})=I_{2L},
\qquad
\operatorname{Cov}(v_{non},v_{non})=I_{D-4L}.
\end{equation}

\textbf{Proof.}
The reference and non-encoding regions are unchanged; hence their coordinates remain independent with unit variance:
\begin{equation}
\operatorname{Cov}(v_{ref},v_{ref})=I_{2L},
\qquad
\operatorname{Cov}(v_{non},v_{non})=I_{D-4L}.
\end{equation}

For the encoding region, since $(x_i,r_i)$ and $(x_j,r_j)$ are independent for $i\neq j$, Theorem 1 implies that $x_i'$ and $x_j'$ are independent, yielding
\begin{equation}
\operatorname{Cov}(x_i',x_j')=0,
\qquad i\neq j.
\end{equation}

For each pair, using Eq.~\ref{eq:compact_encoding} and the independence between $\rho_{x,i}$ and $\phi_{r,i}$,
\begin{equation}
\operatorname{Cov}(x_{i,1}',x_{i,2}')
=
\mathbb{E}[x_{i,1}'x_{i,2}']
=
-(1-2m_i)^2
\mathbb{E}[\rho_{x,i}^2]\,
\mathbb{E}[\sin\phi_{r,i}\cos\phi_{r,i}]
=
0.
\end{equation}

Together with Eq.~\ref{al:variance}, this gives
\begin{equation}
\operatorname{Cov}(x_i',x_i')=I_2,
\end{equation}
and thus
\begin{equation}
\operatorname{Cov}(v'_{enc},v'_{enc})=I_{2L}.
\end{equation}
\hfill

\paragraph{Property 4. Zero Covariance with the Non-encoding Region.}
The non-encoding region is uncorrelated with both the encoding and reference regions:
\begin{equation}
\operatorname{Cov}[v'_{enc},v_{non}]=0,
\qquad
\operatorname{Cov}[v_{ref},v_{non}]=0.
\end{equation}

\textbf{Proof.}
Under $z\sim\mathcal{N}(0,I_D)$, the original encoding, reference, and non-encoding regions are mutually independent. Since the reference and non-encoding regions remain unchanged during embedding, their independence is preserved, yielding
\begin{equation}
\operatorname{Cov}(v_{ref},v_{non})=0.
\end{equation}

For the encoding region, each transformed pair satisfies $x_i'=f_i(x_i,r_i)$, which depends only on $(x_i,r_i)$. Since $(x_i,r_i)$ is independent of $v_{non}$, Theorem 1 implies that $x_i'$ and $v_{non}$ remain independent. Therefore,
\begin{equation}
\operatorname{Cov}(v'_{enc},v_{non})=0.
\end{equation}
\hfill

\paragraph{Property 5. Zero Covariance between Encoding Pairs and Non-paired Reference Pairs.}
For any encoding pair $x_i'$ and any non-paired reference pair $r_j$ with $i\neq j$,
\begin{equation}
\operatorname{Cov}(x_i',r_j)=0.
\end{equation}

\textbf{Proof.}
Since $x_i'=f_i(x_i,r_i)$ depends only on $(x_i,r_i)$, and $r_j$ is independent of $(x_i,r_i)$ for $i\neq j$, Theorem 1 implies that $x_i'$ and $r_j$ are independent. Therefore,
\begin{equation}
\operatorname{Cov}(x_i',r_j)=0,
\qquad i\neq j.
\end{equation}
\hfill

\paragraph{Property 6. Non-zero Covariance between Paired Encoding and Reference Pairs.}
For each encoding pair $x_i'$ and its paired reference pair $r_i$, the cross-covariance is non-zero:
\begin{equation}
\operatorname{Cov}(r_i,x_i')\neq 0.
\end{equation}

\textbf{Proof.}
Let the reference pair be
\begin{equation}
r_i=[r_{i,1},r_{i,2}]
=
[\rho_{r,i}\cos\phi_{r,i},\rho_{r,i}\sin\phi_{r,i}].
\end{equation}

From Eq.~\ref{eq:compact_encoding}, the encoding pair can be written as
\begin{equation}
x_i'
=
\left[
-(1-2m_i)\,\rho_{x,i}\sin\phi_{r,i},
\ (1-2m_i)\,\rho_{x,i}\cos\phi_{r,i}
\right].
\end{equation}

Since all coordinates have zero mean (Eq.~\ref{al:mean}), the cross-covariance matrix is
\begin{equation}
\operatorname{Cov}(r_i,x_i')
=
\mathbb{E}[r_i(x_i')^\top].
\end{equation}

Computing each entry and using the independence between $\rho_{r,i}$, $\rho_{x,i}$, and $\phi_{r,i}$, we obtain
\begin{align}
\mathbb{E}[r_{i,1}x_{i,1}']
&=
-(1-2m_i)\,
\mathbb{E}[\rho_{r,i}\rho_{x,i}]\,
\mathbb{E}[\cos\phi_{r,i}\sin\phi_{r,i}]
=0,\\
\mathbb{E}[r_{i,1}x_{i,2}']
&=
(1-2m_i)\,
\mathbb{E}[\rho_{r,i}\rho_{x,i}]\,
\mathbb{E}[\cos^2\phi_{r,i}]
=
(1-2m_i)\tfrac{1}{2}
\mathbb{E}[\rho_{r,i}\rho_{x,i}],\\
\mathbb{E}[r_{i,2}x_{i,1}']
&=
-(1-2m_i)\,
\mathbb{E}[\rho_{r,i}\rho_{x,i}]\,
\mathbb{E}[\sin^2\phi_{r,i}]
=
-(1-2m_i)\tfrac{1}{2}
\mathbb{E}[\rho_{r,i}\rho_{x,i}],\\
\mathbb{E}[r_{i,2}x_{i,2}']
&=
(1-2m_i)\,
\mathbb{E}[\rho_{r,i}\rho_{x,i}]\,
\mathbb{E}[\sin\phi_{r,i}\cos\phi_{r,i}]
=0.
\end{align}

Thus,
\begin{equation}
\operatorname{Cov}(r_i,x_i')
=
(1-2m_i)
\begin{bmatrix}
0 & \frac{1}{2}\mathbb{E}[\rho_{r,i}\rho_{x,i}]\\
-\frac{1}{2}\mathbb{E}[\rho_{r,i}\rho_{x,i}] & 0
\end{bmatrix}.
\end{equation}

Since $r_i$ and $x_i$ are independent standard Gaussian pairs, their radii are independent Rayleigh random variables with
\begin{equation}
\mathbb{E}[\rho_{r,i}]
=
\mathbb{E}[\rho_{x,i}]
=
\sqrt{\frac{\pi}{2}},
\end{equation}
thus
\begin{equation}
\mathbb{E}[\rho_{r,i}\rho_{x,i}]
=
\mathbb{E}[\rho_{r,i}]
\mathbb{E}[\rho_{x,i}]
=
\frac{\pi}{2}.
\end{equation}

Therefore,
\begin{equation}
\operatorname{Cov}(r_i,x_i')
=
(1-2m_i)
\begin{bmatrix}
0 & \frac{\pi}{4}\\
-\frac{\pi}{4} & 0
\end{bmatrix}
\neq 0.
\end{equation}

Hence, each encoding pair is statistically correlated with its paired reference pair.
\hfill

\paragraph{Summary.}
Combining the above properties, the covariance matrix of $v'$ has the block form
\begin{equation}
\operatorname{Cov}(v')
=
\begin{bmatrix}
I_{2L} & A & 0\\
A^\top & I_{2L} & 0\\
0 & 0 & I_{D-4L}
\end{bmatrix},
\end{equation}
where the block $A=\operatorname{Cov}(v'_{enc},v_{ref})$ is block diagonal:
\begin{equation}
A=
\operatorname{diag}(A_1,A_2,\dots,A_L),
\end{equation}
with
\begin{equation}
A_i
=
\operatorname{Cov}(x_i',r_i)
=
(1-2m_i)
\begin{bmatrix}
0 & -\frac{\pi}{4}\\
\frac{\pi}{4} & 0
\end{bmatrix}.
\end{equation}

\section{Magnitude-Driven Robustness}
\label{sec:appendix_proof1}

To justify selecting high-magnitude latent pairs for watermark embedding, we show that larger magnitudes lead to smaller angular deviations under perturbations.

\paragraph{Property 7. Angular Error Variance.}
Let $x\in \mathbb{R}^2$ be a latent pair with magnitude $\rho=\|x\|_2$. Suppose it is perturbed by isotropic Gaussian noise $\epsilon\sim\mathcal{N}(0,\sigma^2 I_2)$. Under small perturbations ($\|\epsilon\|\ll\rho$), the induced angular error $\Delta\phi$ satisfies
\begin{equation}
\operatorname{Var}(\Delta\phi)\approx \frac{\sigma^2}{\rho^2}.
\end{equation}

\textbf{Proof.}
Let $\hat{x}=x+\epsilon$, and decompose the perturbation as
\begin{equation}
\epsilon=\epsilon_{\parallel}+\epsilon_{\perp},
\end{equation}
where $\epsilon_{\parallel}$ and $\epsilon_{\perp}$ denote the radial and tangential components, respectively. Under sufficiently small perturbations ($\|\epsilon\|\ll\rho$), the tangential displacement approximately equals the arc length induced by the angular deviation:
\begin{equation}
\|\epsilon_{\perp}\|\approx \rho\,\Delta\phi,
\end{equation}
which gives
\begin{equation}
\Delta\phi\approx \frac{\epsilon_{\perp}}{\rho}.
\end{equation}

Since $\epsilon\sim\mathcal{N}(0,\sigma^2 I_2)$ is isotropic, its projection onto any direction remains Gaussian, i.e.,
\begin{equation}
\epsilon_{\perp}\sim\mathcal{N}(0,\sigma^2).
\end{equation}
Therefore,
\begin{equation}
\operatorname{Var}(\Delta\phi)
\approx
\frac{1}{\rho^2}\operatorname{Var}(\epsilon_{\perp})
=
\frac{\sigma^2}{\rho^2}.
\end{equation}
\hfill

\section{Gaussianity Analysis of Gaussian Shading}
\label{sec:appendix_gs_analysis}

\subsection{Statistical Formulation}

Gaussian Shading\citep{Yang_2024_CVPR} embeds watermark information into latent space through distribution-preserving sampling. We follow the
original formulation and model the process in three steps: watermark diffusion, randomization via a secret key, and interval-based sampling.

Let $s \in \{0,1\}^k$ denote the original watermark. After spatial diffusion, we obtain a diffused watermark $s_d$. Then, a stream cipher with secret key $K$ is applied:
\begin{equation}
m = E(K, s_d),
\end{equation}
where $E(\cdot)$ denotes a cryptographically secure stream cipher. As a result, $m$ is a pseudorandom bit sequence,
which is computationally indistinguishable from a uniformly random sequence.

Next, the randomized bit sequence $m$ is used to drive sampling.
Each $l$-bit segment of $m$ is interpreted as an integer
\begin{equation}
y \in \{0,1,\dots,2^l-1\}.
\end{equation}

Let $f(x)$ denote the density of the standard Gaussian distribution $\mathcal{N}(0,1)$, and let $\mathrm{ppf}(\cdot)$ be its quantile function. The real line is partitioned into $2^l$ equal-probability intervals:
\begin{equation}
\left[\mathrm{ppf}\left(\frac{i}{2^l}\right),
\mathrm{ppf}\left(\frac{i+1}{2^l}\right)\right], \quad i=0,\dots,2^l-1.
\end{equation}

Given $y=i$, the latent coordinate is sampled as:
\begin{equation}
z_i = \mathrm{ppf}\left(\frac{u+i}{2^l}\right), \quad u \sim \mathcal{U}(0,1).
\end{equation}

\paragraph{Special case: sign-controlled sampling.}
When $l=1$, the Gaussian distribution is partitioned into two
equal-probability halves. In this case, the above formulation reduces
to a sign-controlled sampling scheme. Specifically, defining
\begin{equation}
S_i = 2y - 1 \in \{-1,+1\},  y \in \{0,1\},
\end{equation}
we can rewrite the sampling process as:
\begin{equation}
z_i = S_i R_i,
\end{equation}
where $R_i = |G_i|$ and $G_i \sim \mathcal{N}(0,1)$.

\subsection{Mean, Variance, and Covariance}

We analyze two key-management settings: a fixed-key setting and a flexible key setting. In both cases, the user-specific watermark $s$ is fixed for the same user, and its diffused version $s_d$ is also fixed.

We adopt the sign-controlled formulation introduced above:
\begin{equation}
z_i = S_i R_i,
\end{equation}
where $R_i = |G_i|$ and $G_i \sim \mathcal{N}(0,1)$. Thus,
\begin{equation}
\mathbb{E}[R_i] = \sqrt{\frac{2}{\pi}},
\qquad
\mathbb{E}[R_i^2] = 1.
\end{equation}

\subsubsection{Fixed-key}

If the same key $K$ is reused for all generated images of the same user, then the randomized sequence $m = E(K, s_d)$ is fixed. Consequently, the sign pattern $S_i$ is fixed for each coordinate.

The latent coordinate becomes
\begin{equation}
z_i = S_i R_i.
\end{equation}

The mean is
\begin{equation}
\mathbb{E}[z_i \mid S_i]
=
S_i \mathbb{E}[R_i]
=
S_i \sqrt{\frac{2}{\pi}}.
\end{equation}

The second moment is
\begin{equation}
\mathbb{E}[z_i^2 \mid S_i]
=
\mathbb{E}[S_i^2 R_i^2]
=
1.
\end{equation}

Thus, the variance is
\begin{equation}
\operatorname{Var}(z_i \mid S_i)
=
1 - \frac{2}{\pi}.
\end{equation}

For two different coordinates $i \neq j$,
\begin{equation}
z_i = S_i R_i,
\qquad
z_j = S_j R_j.
\end{equation}

Since $R_i$ and $R_j$ are independently sampled,
\begin{equation}
\mathbb{E}[z_i z_j \mid S_i, S_j]
=
S_i S_j \mathbb{E}[R_i]\mathbb{E}[R_j]
=
S_i S_j \frac{2}{\pi}.
\end{equation}

Also,
\begin{equation}
\mathbb{E}[z_i \mid S_i] \mathbb{E}[z_j \mid S_j]
=
S_i S_j \frac{2}{\pi}.
\end{equation}

Therefore,
\begin{equation}
\operatorname{Cov}(z_i, z_j \mid S_i, S_j) = 0,
\qquad i \neq j.
\end{equation}

Hence, under a fixed key, different coordinates remain uncorrelated, but the variance is reduced:
\begin{equation}
\operatorname{Cov}(z \mid S)
=
\left(1 - \frac{2}{\pi}\right) I.
\end{equation}

\subsubsection{Flexible Key}

If a flexible key $K$ is sampled for each generated image, then the
randomized sequence $m = E(K, s_d)$ varies across images. Under the stream-cipher assumption, the induced sign variables satisfy
\begin{equation}
\Pr(S_i = +1) = \Pr(S_i = -1) = \frac{1}{2}.
\end{equation}

Assuming independence between $S_i$ and $R_i$, we have
\begin{equation}
\mathbb{E}[z_i]
=
\mathbb{E}[S_i R_i]
=
\mathbb{E}[S_i]\mathbb{E}[R_i]
=
0.
\end{equation}

The second moment is
\begin{equation}
\mathbb{E}[z_i^2]
=
\mathbb{E}[S_i^2 R_i^2]
=
1.
\end{equation}

Thus,
\begin{equation}
\operatorname{Var}(z_i) = 1.
\end{equation}

For $i \neq j$, assume that $S_i$ and $S_j$ are independent, $R_i$ and $R_j$ are independent, and $(S_i,S_j)$ is independent of $(R_i,R_j)$. Then
\begin{equation}
\mathbb{E}[z_i z_j]
=
\mathbb{E}[S_i S_j R_i R_j]
=
\mathbb{E}[S_i]\mathbb{E}[S_j]\mathbb{E}[R_i]\mathbb{E}[R_j]
=
0.
\end{equation}

Since $\mathbb{E}[z_i]=\mathbb{E}[z_j]=0$, we have
\begin{equation}
\operatorname{Cov}(z_i, z_j) = 0,
\qquad i \neq j.
\end{equation}

Therefore,
\begin{equation}
\operatorname{Cov}(z) = I.
\end{equation}

\subsubsection{Summary}

\paragraph{Fixed key.}
When the same key is reused, the sign pattern $S_i$ is fixed, and we have
\begin{equation}
\mathbb{E}[z_i]=S_i\sqrt{\frac{2}{\pi}}, \quad
\operatorname{Var}(z_i)=1-\frac{2}{\pi}, \quad
\operatorname{Cov}(z)=\left(1-\frac{2}{\pi}\right)I.
\end{equation}
This deviation from the standard Gaussian prior leads to noticeable degradation in generation quality.

\paragraph{Flexible Key.}
When a flexible key is used for each image, the induced signs $S_i$ are balanced and approximately independent, yielding
\begin{equation}
\mathbb{E}[z_i]=0, \quad \operatorname{Var}(z_i)=1, \quad \operatorname{Cov}(z)=I.
\end{equation}
Thus, the i.i.d. Gaussian prior is preserved. However, this requires storing a unique key for each image, introducing substantial key storage and management overhead.

\section{Pseudocode}

\subsection{Latent Angular Watermarking}

\begin{algorithm}[t]
\caption{Latent Angular Watermarking (LAW)}
\label{alg:angular_watermarking}
\begin{algorithmic}[1]
\Require Latent tensor $z$, watermark message $m \in \{0,1\}^L$, corrupted image $\hat{I}$
\Ensure Watermarked latent $z_m$ (Encoding) / decoded watermark message $\hat{m}$ (Decoding)

\Statex \textbf{// --- Encoding Procedure ---}
\State $v \gets \operatorname{flatten}(z)$
\State Partition $v$ into encoding region $\{x_i\}_{i=1}^L$, reference region $\{r_i\}_{i=1}^L$, and non-encoding region

\For{$i=1$ \textbf{to} $L$}
    \State $\rho_{x,i}\gets \|x_i\|_2$, \quad $\phi_{r,i}\gets \operatorname{atan2}(r_{i,2},r_{i,1})$
    
    \State \textbf{if} $m_i=0$ \textbf{then} $\phi'_{x,i}\gets\phi_{r,i}+\frac{\pi}{2}$ \textbf{else} $\phi'_{x,i}\gets\phi_{r,i}-\frac{\pi}{2}$
    
    \State $x'_i\gets
    \left[
    \rho_{x,i}\cos\phi'_{x,i},
    \rho_{x,i}\sin\phi'_{x,i}
    \right]$
\EndFor

\State Replace $\{x_i\}_{i=1}^L$ with $\{x'_i\}_{i=1}^L$ to obtain $v'=[x'_1,\ldots,x'_L,r_1,\ldots,r_L,\ldots]$
\State \Return $z_m=\operatorname{reshape}(v')$

\Statex \textbf{// --- Decoding Procedure ---}
\State $\hat{z}_T\gets \operatorname{DDIM\_Inversion}(\hat{I})$
\State $\hat{v}\gets \operatorname{flatten}(\hat{z}_T)$
\State Partition $\hat{v}$ into encoding region $\{\hat{x}_i\}_{i=1}^L$, reference region $\{\hat{r}_i\}_{i=1}^L$ and non-encoding region

\For{$i=1$ \textbf{to} $L$}
    \State $\hat{\phi}_{x,i}\gets \operatorname{atan2}(\hat{x}_{i,2},\hat{x}_{i,1})$
    \State $\hat{\phi}_{r,i}\gets \operatorname{atan2}(\hat{r}_{i,2},\hat{r}_{i,1})$
    
    \State $\hat{m}_i\gets
    \begin{cases}
    0, & \textbf{if } \hat{\phi}_{x,i}>\hat{\phi}_{r,i},\\
    1, & \textbf{otherwise}
    \end{cases}$
\EndFor

\State \Return $\hat{m}$
\end{algorithmic}
\end{algorithm}

\subsection{Magnitude-Driven Latent Angular Watermarking}

\begin{algorithm}[t]
\caption{Magnitude-Driven Latent Angular Watermarking (LAW-M)}
\label{alg:pair_first_sort_later}
\begin{algorithmic}[1]
\Require Latent tensor $z$, watermark message $m \in \{0,1\}^L$, corrupted image $\hat{I}$, private key $\mathcal{K}$
\Ensure Watermarked latent $z_m$ (Encoding) / decoded watermark message $\hat{m}$ (Decoding)

\Statex \textbf{// --- Encoding Procedure ---}
\State $v \gets \operatorname{flatten}(z)$
\State Partition $v$ into $D/2$ disjoint pairs 
\State Sort all pairs by magnitude and obtain
$
v_M=[x_1,\dots,x_L,r_1,\dots,r_L,\dots]
$
such that
$
\rho_{x,1}\ge\cdots\ge\rho_{x,L}\ge\rho_{r,1}\ge\cdots\ge\rho_{r,L}
$
\State Store the original indices of $\{x_i\}_{i=1}^L$ and $\{r_i\}_{i=1}^L$ as private key $\mathcal{K}$

\For{$i=1$ \textbf{to} $L$}
    \State $\rho_{x,i}\gets \|x_i\|_2$, \quad $\phi_{r,i}\gets \operatorname{atan2}(r_{i,2},r_{i,1})$
    
    \State \textbf{if} $m_i=0$ \textbf{then} $\phi'_{x,i}\gets\phi_{r,i}+\frac{\pi}{2}$ \textbf{else} $\phi'_{x,i}\gets\phi_{r,i}-\frac{\pi}{2}$
    
    \State $
    x'_i\gets
    \left[
    \rho_{x,i}\cos\phi'_{x,i},
    \rho_{x,i}\sin\phi'_{x,i}
    \right]
    $
\EndFor

\State Replace $x_i$ with $x_i'$ for $i=1,\dots,L$, then restore all pairs to their original positions using the private key $\mathcal{K}$ to obtain $v'$.
\State \Return $z_m=\operatorname{reshape}(v')$, private key $\mathcal{K}$

\Statex \textbf{// --- Decoding Procedure ---}
\State $\hat{z}_T\gets \operatorname{DDIM\_Inversion}(\hat{I})$
\State $\hat{v}\gets \operatorname{flatten}(\hat{z}_T)$
\State Partition $\hat{v}$ into $D/2$ disjoint pairs
\State Recover $\hat{x}_i$ and $\hat{r}_i$ using the private key $\mathcal{K}$ for $i=1,\dots,L$

\For{$i=1$ \textbf{to} $L$}
    \State $\hat{\phi}_{x,i}\gets \operatorname{atan2}(\hat{x}_{i,2},\hat{x}_{i,1})$, \quad
    $\hat{\phi}_{r,i}\gets \operatorname{atan2}(\hat{r}_{i,2},\hat{r}_{i,1})$
    
    \State $
    \hat{m}_i\gets
    \begin{cases}
    0, & \hat{\phi}_{x,i}>\hat{\phi}_{r,i},\\
    1, & \text{otherwise}
    \end{cases}
    $
\EndFor

\State \Return $\hat{m}$
\end{algorithmic}
\end{algorithm}



\section{Experimental settings}
\label{sec:app_settings}

\subsection{Computational Indistinguishability}
\label{sec:appendix_exp_details}

For each watermarking baseline, we curate a dataset based on captions from the MS COCO 2018 training and validation splits. We generate 800 image pairs (watermarked and non-watermarked) for training and 200 pairs for validation. For post-processing methods, the non-watermarked images are original input images. For in-generation methods, the non-watermarked images are defined as images synthesized by the vanilla Stable Diffusion v2.1 using identical text prompts and the corresponding initial Gaussian noise.

For image-level binary classification, we utilize a ResNet-18 model trained for 100 epochs with a learning rate of $10^{-3}$ and a batch size of 64. For latent-level binary classification, a two-layer Multi-Layer Perceptron (MLP) is implemented, undergoing 200 training epochs at a learning rate of $10^{-4}$ with a batch size of 64. All experiments are implemented using PyTorch and conducted on a NVIDIA RTX  5090 GPU.


\subsection{Covariance Matrix Analysis}
\label{sec:covariance}

Existing Gaussian-mapping watermarking include Gaussian Shading~\citep{Yang_2024_CVPR} and PRC~\citep{gunn_2025_undetectable}. Since the covariance matrix characterizes the internal correlation, we compare the covariance matrices produced by our method and these two approaches.

Specifically, we generate samples by embedding a 2-bit watermark into a 16-dimensional Gaussian vector. We independently sample $N=10{,}000$ such vectors, and stack them column-wise to form
\begin{equation}
T =
\begin{bmatrix}
\mathbf{z}^{(1)} & \mathbf{z}^{(2)} & \cdots & \mathbf{z}^{(N)}
\end{bmatrix}
\in \mathbb{R}^{16 \times N},
\end{equation}
where each column $\mathbf{z}^{(i)} \in \mathbb{R}^{16}$ denotes one watermarked latent sample. The covariance matrix is then estimated as
\begin{equation}
\Sigma = \frac{1}{N} T T^\top .
\end{equation}

\section{Impact of Latent Distribution Shifts on Generation Quality and DDIM Inversion}
\label{sec:appendix_cov_analysis}

Diffusion models are typically trained under the standard Gaussian prior $z_T \sim \mathcal{N}(0,I)$. If the mean or variance of the initial latent deviates from this prior, i.e., $\mathbb{E}[z_T]\neq 0$ or $\mathrm{Var}(z_T)\neq I$, the input
noise distribution no longer matches the training distribution. A mean shift introduces a systematic bias in the denoising process, while a variance shift changes the effective noise scale and thus causes a mismatch with the scheduled
inverse diffusion process. As a result, the UNet noise prediction becomes biased at each denoising step, leading to accumulated errors and degraded generation quality.

More importantly, if the covariance matrix deviates from the identity matrix, i.e., $\mathrm{Cov}(z_T)=\Sigma \neq I$, the latent is no longer isotropic. Geometrically, the noise distribution changes from a spherical Gaussian
to an ellipsoidal Gaussian, introducing correlations among latent dimensions. Since the UNet is trained to denoise isotropic Gaussian noise, such correlation structure causes direction-dependent prediction errors. This further misaligns the inverse diffusion process with the DDIM inversion, resulting in larger inversion errors.

To further investigate the impact of covariance structure on inversion accuracy, we conduct a controlled experiment by introducing correlations into the initial Gaussian latent. Specifically, we construct a correlated Gaussian latent as
\begin{equation}
z = \sqrt{\rho}\, z_{\text{shared}} + \sqrt{1-\rho}\, z_{\text{private}},
\end{equation}
where $z_{\text{shared}}$ is a shared Gaussian across dimensions and $z_{\text{private}}$ is independently sampled noise. This construction induces non-zero correlations among latent dimensions, resulting in a covariance matrix $\Sigma \neq I$.

We evaluate inversion accuracy by measuring the $\ell_1$ distance between the reconstructed latent and the original latent. The results show that the average $\ell_1$ distance increases significantly from $0.3304$ (independent case) to $0.9274$ (correlated case). This substantial degradation demonstrates that introducing covariance correlations leads to a clear mismatch between the inverse diffusion process and the inversion DDIM, thereby increasing inversion error.

\section{Post-processing and Regeneration Attack Settings}
\label{sec:pr_settings}

\textbf{Post-processing Attacks.}
We evaluate robustness under a range of common post-processing attacks. The parameter ranges are defined as follows:
\begin{itemize}
    \item \textbf{PNG:} $[0]$
    \item \textbf{Gaussian noise:} $[0.05, 0.10, 0.15, 0.20, 0.25, 0.30]$
    \item \textbf{Brightness:} $[1, 2, 3, 4, 5, 6, 7, 8]$
    \item \textbf{Random drop:} $[0.1, 0.2, 0.3, 0.4, 0.5, 0.6, 0.7]$
    \item \textbf{Gaussian blur:} $[5, 10, 15, 20, 25, 30]$
    \item \textbf{JPEG compression:} $[90, 80, 70, 60, 50, 40, 30, 20, 10]$
    \item \textbf{Median filter:} $[4, 6, 8, 10, 12, 14, 16]$
    \item \textbf{Resize:} $[0.7, 0.6, 0.5, 0.4, 0.3, 0.2, 0.1]$
\end{itemize}

\textbf{Regeneration Attacks.}
We further evaluate robustness under a range of regeneration attacks. The parameter settings are defined as follows:
\begin{itemize}
    \item \textbf{Diffusion-based regeneration:} We adopt the Stable Diffusion 2.1 and evaluate robustness under different noise addition steps $t \in \{10, 20, 30, 50, 80, 100, 150, 200\}$.
    \item \textbf{VAE-based regeneration:} We employ two pre-trained VAE models, denoted as Regen-VAE-B\citep{ballé2018variational} and Regen-VAE-C\citep{Cheng_2020_CVPR}. The robustness is evaluated across quality factor $q \in \{1, 2, 3, 4, 5, 6\}$.
\end{itemize}

\begin{figure}[t]
    \centering
    \includegraphics[width=\textwidth]{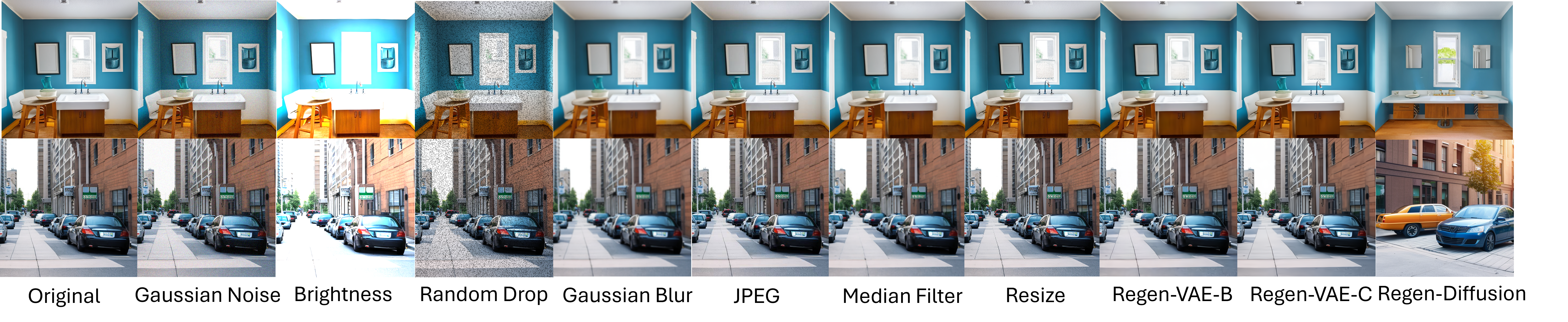}
    \caption{Qualitative comparison of image quality across various attack methods.}
    \label{fig:fixed_attacks}
\end{figure}

The Figure \ref{fig:fixed_attacks} presents examples of images subjected to various attacks, including Gaussian noise ($\sigma=0.05$), brightness adjustment (factor = 2), random drop (ratio = 0.1), Gaussian blur (kernel size = 9), JPEG compression (quality = 70), median filtering (kernel size = 5), resizing (ratio = 0.5), VAE-based regeneration (Regen-VAE-B/C, quality factor = 4), and diffusion-based regeneration (t=100).

\textbf{EBRA\citep{liu2023erase}.}
We apply a lattice-based perturbation by periodically selecting one pixel every $q$ pixels along both spatial dimensions and replacing the selected pixel values with random intensities across all channels. By default, we set $q=5$.

\end{document}